# Significance of parallel computing on the performance of Digital Image Correlation algorithms in MATLAB

Andreas Thoma[1,*] and Sridhar Ravi[2]

[1] Royal Melbourne Institute of Technology (RMIT University), Melbourne, Australia and FH Aachen, Aachen, Germany
[2] Royal Melbourne Institute of Technology (RMIT University), Melbourne, Australia

* Corresponding author.
E-mail address: a.thoma@fh-aachen.de (Andreas Thoma)

## Abstract

**Digital Image Correlation (DIC) is a powerful tool used to evaluate displacements and deformations in a non-intrusive manner. By comparing two images, one of the undeformed reference state of a specimen and another of the deformed target state, the relative displacement between those two states is determined. DIC is well known and often used for post-processing analysis of in-plane displacements and deformation of specimen. Increasing the analysis speed to enable real-time DIC analysis will be beneficial and extend the field of use of this technique. Here we tested several combinations of the most common DIC methods in combination with different parallelization approaches in MATLAB and evaluated their performance to determine whether real-time analysis is possible with these methods. To reflect improvements in computing technology different hardware settings were also analysed. We found that implementation problems can reduce the efficiency of a theoretically superior algorithm such that it becomes practically slower than a suboptimal algorithm. The Newton-Raphson algorithm in combination with a modified Particle Swarm algorithm in parallel image computation was found to be most effective. This is contrary to theory, suggesting that the inverse-compositional Gauss-Newton algorithm is superior. As expected, the Brute Force Search algorithm is the least effective method. We also found that the correct choice of parallelization tasks is crucial to achieve improvements in computing speed. A poorly chosen parallelisation approach with high parallel overhead leads to inferior performance. Finally, irrespective of the computing mode the correct choice of combinations of integer-pixel and sub-pixel search algorithms is decisive for an efficient analysis. Using currently available hardware real-time analysis at high framerates remains an aspiration.**

## 1 Introduction

In many different engineering applications the measurement of displacements and deformations play an important role [1]. Due to limitations of invasive methods for strain estimations, there exists a strong motivation for the development of contactless measurement techniques. Optical based strain measurement systems have a special importance in the field of measurement systems as they offer a potential, non-invasive, high throughput strain measurement system. Therefore, over the last decade several different measurement systems for optical displacement measurement have been developed [2, 3]. Digital Image Correlation (DIC) gained a lot of popularity not only because of its simplicity and accuracy [4] but also because of its robustness and wide field of applications [5]. After this method was first published by Peters and Ranson [6] in the early 1980s, the DIC method has been continuously improved by different researchers, while the basic principle remained the same. Firstly, accurately time resolved images of a specimen undergoing deformation are acquired. The subsequent analysis is conducted between image pairs separated over time where one of the images is considered as the reference state while the other represents the deformed state of the specimen. Finally, the image pairs are compared by dividing the reference image into sub-images and searching for the same sub-image in the image of the deformed state. By estimating the inter-image positions of the sub-image pairs, a local displacement is determined.

The method of dividing the image into sub-images, searching between images and processing routines have undergone many changes and improvements over the years to develop accurate techniques for use in a variety of engineering fields [7]. However, one of the main disadvantages of DIC is its high computational burden and the relatively slow image processing time, which imposes significant limitations on the frame rate of image acquisition, and post-processing of images. Additionally, DIC is usually restricted to 2D analyses because of its high processing times [5].

Here we tested the most common computational strategies used in DIC and evaluated their performance under serial and parallel computational implementation in MATLAB. MATLAB is a computing environment commonly used by engineers, researchers and economists [8]. MATLAB comes with a variety of powerful toolboxes and functions, allowing comparably easy use of higher-level functionalities and integration of complex functions into self-developed programs. However, MATLAB is slower than other programming languages and is limited in control of low level functionalities as for example memory control [9, 10]. However, because of its widespread use, we decided to focus on MATLAB.

We sought to combine the most efficient integer and sub pixel displacement DIC methods such that they advance each other. A highly modular program was developed to investigate advantages and disadvantages of different computational methods and their combinations. These methods were tested, compared and improved until the most efficient and optimised combination of methods was found. Two different parallelization approaches were adopted, on the one hand, sub-image parallelisation in which sub-images are processed in parallel and on the other hand, image parallelisation in which whole image pairs are processed in parallel. Besides quantifying the influences of these top-level parallelisation approaches, the influence hardware settings on the overall efficiency of DIC computations was also evaluated.

## 2 Theoretical background

The basic idea of DIC is to determine displacements between two states of an object of interest by comparing two images of this object in the two different states; a reference image, e.g. a specimen in an undeformed state, and a target image, e.g. a specimen in a deformed state. For this purpose two functions, one representing the gray level in the reference image f(x, y) and one for the target image g(x*,y*) at the positions [x,y] and [x*,y*], respectively, are defined. The reference image is divided into several rectangular sub-images. Each sub-image has a specific gray level distribution, which is compared to a sub-image of the same size within the target image. This comparison is done by correlation coefficients, which provide a measure of similarity. The strategy of DIC methods can be roughly divided into two steps: first, the integer-pixel displacement search, which determines the displacement of a sub-image as an integer value. Second, a sub-pixel displacement search algorithm capable of determining sub-pixel displacements. However, in order to reduce computational time generally the integer-pixel search is preformed to provide initial guesses to initiate the sub-pixel displacement algorithm.

$$C(p) = \frac{\sum_{x=-M}^{M}\sum_{y=-M}^{M}[f(x,y) - \bar{f}] \times [g(x^*,y^*) - \bar{g}]}{\sqrt{\sum_{x=-M}^{M}\sum_{y=-M}^{M}[f(x,y) - \bar{f}]^2} \sqrt{\sum_{x=-M}^{M}\sum_{y=-M}^{M}[g(x^*,y^*) - \bar{g}]^2}} \quad (1)$$

### 2.1 Integer pixel displacement

The integer pixel displacement can be determined by a variety of algorithms. The available algorithms differ in many aspects but all of them calculate some kind of correlation coefficient. Currently, a zero-normalized cross-correlation coefficient, as presented in equation (1), is used most commonly [11].

Here, f(x,y) and g(x*,y*) represent the gray levels at a certain position in reference and target image, while $\bar{f}$ and $\bar{g}$ represent the average gray levels of the sub-images with a size of (2M+1)x(2M+1) pixels. This correlation coefficient C of a point **p** = $(x,y)^T$ reaches its maximum if the gray level distributions of both sub-images are equal. This implies that the displacement is determined by determining the position at which the correlation coefficient is maximal.

With this knowledge, the displacement can be determined by a simple but time-consuming Brute Force Search (BFS) algorithm to allocate the position with the highest correlation coefficient. A Brute Force Search algorithm evaluates all possible positions and determines the position with the highest correlation coefficient [12].

Another, smarter approach is a Particle Swarm Optimization. In this optimization strategy particles are initialised randomly at position **p** = ($x_1$,$x_2$) with velocity **v** = ($v_1$,$v_2$). Through iteration with different generations "t" of particle "i" an optimal solution is found. The next generation is determined according to:

$$v_{id}(t+1) = wv_{id}(t) + c_1 r_1 [p_{best\ id} - p_{id}(t)] + c_2 r_2 [g_{best\ d} - p_{id}(t)] \quad (2)$$

$$p_{id}(t+1) = p_{id}(t) + v_{id}(t+1) \quad (3)$$

Where d represents the directions x and y. $c_1$ and $c_2$ are predefined acceleration coefficients which influence cognitive, e.g. local, optimization behaviour and social, e.g. global, optimization behaviour, respectively. The parameters $r_1$ and $r_2$ are independent random values between zero and one.

Each particle has a correlation coefficient and a history with one best position **p**$_{best}$ at which it has perceived its maximal correlation coefficient $c_{best}$ so far. One particle has perceived the highest correlation coefficient $c_{best}$ of the whole swarm at the global best position $g_{best}$.

The generation dependent weight factor $w$ is calculated according to:

$$w(t) = 0{,}9 - \frac{t}{2G_{max}} \quad (4)$$

With the maximum number of generations $G_{max}$.

According to Wu, et al. [13] a maximum of 5 generations and a correlation coefficient of 0.75 are sufficient for efficient DIC calculations.

Because these algorithms shall be used in real-time, small deformations between two consecutive images can be assumed. Therefore, the integer-pixel search algorithm looks for the target sub-image in an area of 25 pixels around the reference sub-image.

## 2.2 Sub-pixel displacement

Limiting the displacements of the sub-image to an integer value leads to an accuracy of ±0.5 pixel. To further increase accuracy an interpolation approach have been introduced to determine the displacement of the sub-image between the two states to sub-pixel values [13]. In this approach, it is assumed that a pixels gray level equals the gray level at its centre. The most basic approach, bilinear interpolation, to determine the gray level at an arbitrary position G(x*, y*) is as following:

$$G(x^*, y^*) = a_{00} + a_{10}x' + a_{01}y' + a_{11}x'y' \quad (5)$$

G(x*, y*) denotes the gray level distribution at an arbitrary sub-pixel position (x*, y*), while x' and y' denote the distance along the x and y axis from the next integer pixel position to (x*, y*) while $a_{00}$, $a_{01}$, $a_{10}$ and $a_{11}$ are the coefficients of the bilinear interpolation function. By means of a linear equation system those coefficients can be determined by the gray levels of surrounding integer pixel positions. Besides this simple bilinear approach, other methods such as bicubic interpolation [14, 15] or higher order spline interpolation [16], have been used as well. Generally, the higher the approaches' order the higher the accuracy at the expense of computational effort [4].

Unfortunately, bilinear interpolation just gives gray levels at certain positions within four pixels. To find the position with the highest correlation also a search algorithm is required. Besides the well known Newton Raphson (NR) algorithm a more efficient Inverse-Compositional Gauss-Newton (IC-GN) approach is commonly used [17]. The IC-GN algorithm uses an affine warp function together with an interpolation approach to backwardly calculate the position of the reference sub-image. Contrary to the NR algorithm, in the IC-GN several parameters can be precomputed and it is not necessary to update them in every iteration step. A detailed explanation can be found in a publication by Pan, Li and Tong [18]. Finally, usage of look-up tables for calculation results required multiple times reduce the computational effort further [18].

## 2.3 Parallel computation

Parallel computation is one of the most commonly used methods to reduce computational time. Therefore, it plays an important role in the development of new programs and the optimization of old ones and significant research concerning this topic is ongoing [19, 20]. Despite its advances and benefits, parallel computation also has its own challenges and is a very complex field with several limitations and restrictions [20]. Nonetheless, some studies have been published recently, concerning parallel computing in DIC, either for CPU computation [13, 17], GPU computation [21] or a mixture of both [22].

Parallel computing splits up one main task into several sub-tasks which are performed simultaneously. Unfortunately, this is only possible if the sub-tasks can be processed separately. In DIC the search for the target sub-images position of different reference sub-images can be processed in parallel within some approaches because they are completely independent of each other. Additionally, whole pairs of reference image and target image can be processed simultaneously on different workers. Parallel computation can be used for both, integer-pixel displacement and sub-pixel displacement search algorithms.

Due to the vastness and complexities of parallel computation procedures, this study has limited itself to a discussion about the same in general terms while assessment of the different DIC algorithms and the significance of parallel computations will be conducted on MATLAB, one of the most commonly used platforms for DIC implementation.

**GPU parallel computation in MATLAB**

Parallel computation splits up one task to perform it on several workers simultaneously. To do so it is necessary to handle data and information such that they are distributed correctly and transferred to the assigned worker. For parallel processing using the GPU in MATLAB, NVIDIA's CUDA GPU computation technology is utilized [10]. CUDA is a programming technique developed by NVIDIA which shifts program parts to the GPU [23]. The CUDA technology is included into MATLAB as a simple to use toolkit. MATLAB introduced a new data type called "*gpuArray*". This data type is similar to a usual array but has, additionally included the information that all tasks related to this array have to be performed on the GPU. More than 100 MATLAB functions support this data type and shift, via the CUDA technology, the computational task to the GPU if casted with a *gpuArray* [10]. This technology's bottleneck lies within the data transfer. For every computational task all the required data has to be transported from the CPU to the GPU before it can be processed. Afterwards any results have to be transferred back from the GPU to the CPU [10]. Because the GPU is attached to the CPU via a PCI express bus the data transfer is slower than for standard CPU usage [24]. Summarizing, the computational speed is heavily limited by the required amount of data transfer.

**CPU parallel computation in MATLAB**

Besides GPU parallel computation MATLAB offers the possibility to perform tasks in parallel on different CPU workers [10]. In order to enable parallel computation, MATLAB creates a parallel computation pool of different workers of a multicore CPU and processes different independent tasks in parallel [10].

MATLAB provides the user with two different possibilities to execute computing scripts in CPU parallel computation [10]. On the one hand, the "parfor" loop function, which works similar to a standard "for" loop but executes its command section in parallel. On the other hand, the "parfeval" function, which allows asynchronous sub-function call and processing on a different worker without stopping the main function to wait for any results of the called sub-function [10].

While CPU parallel processing may seem advantageous, data transfer possess a bottleneck here as well. For CPU parallel computation this is not as critical as for GPU parallel computation but still an important factor which must be addressed [10]. MATLAB creates a copy of the variable for every worker and passes it to the worker. If a large volume of data has to be handled by every worker substantial time is required to copy and transfer it. Therefore, it may be possible that the benefits of the parallel computation are diminished by the time spent for organising and transferring data.

# 3 Tests and comparisons

Here, we developed a standalone DIC program that uses a combination of different DIC methods including leveraging the benefits of parallel computing to perform DIC analyses as efficiently as possible. The most commonly used and most efficient algorithms were programmed in MATLAB as shown in Table 1&2. Some modifications of those algorithms as well as different versions were integrated, tested and evaluated. The testing platform we developed is highly modular such that it is possible to choose any combination of mathematical search algorithms. To evaluate the performance of different parallelization approaches two different ways of parallelizing the image evaluation, parallel im-age processing and parallel sub-image processing, are implemented. All methods and parallelisation approaches are implemented completely independent, to enable an optimised program structure for every module.

| Integer-pixel search algorithms |
|---|
| Brute Force Search Algorithm |
| Particle Swarm Optimization |
| Modified Particle Swarm Optimization (integrated Star Search algorithm) |

Table 1: Overview of Integer-pixel search algorithms

| Sub-pixel search algorithms |
|---|
| Newton Raphson Method |
| Inverse-Compositional Gauss Newton Method self-implemented |
| Inverse-Compositional Gauss Newton Method by Baker and Matthews [25] |

Table 2: Overview of Sub-Pixel search algorithms

The two standard DIC search algorithms, Brute Force Search algorithm and Newton Raphson method, are implemented as reference methods. Additionally, one of the two Particle Swarm Optimization implemented here is modified such that a gradient descent search algorithm, the star search algorithm, is integrated within the Particle Swarm Optimization algorithm. Furthermore, two versions of the Inverse-Compositional Gauss Newton method are implemented. One variation was completely implemented by the authors and another one taken from a publication by Baker and Matthews [25], which was then integrated in the developed software. The two implementations of the Inverse-Compositional Gauss-Newton algorithm differ in details as the overall program structure and usage of higher-level MATLAB functions.

Finally, the accuracy and efficiency of combinations of different methods is tested with two different sets of test data. The first data set consists of 204 images of a rectangular specimen undergoing deformation due to tensile forces, taken in the RMIT Materials Laboratory. The second set consists of eleven images showing different states of a deformation process taken from Blaber, et al. [26]. A sample of three consecutive images of each of the two data sets is presented in Fig 1.

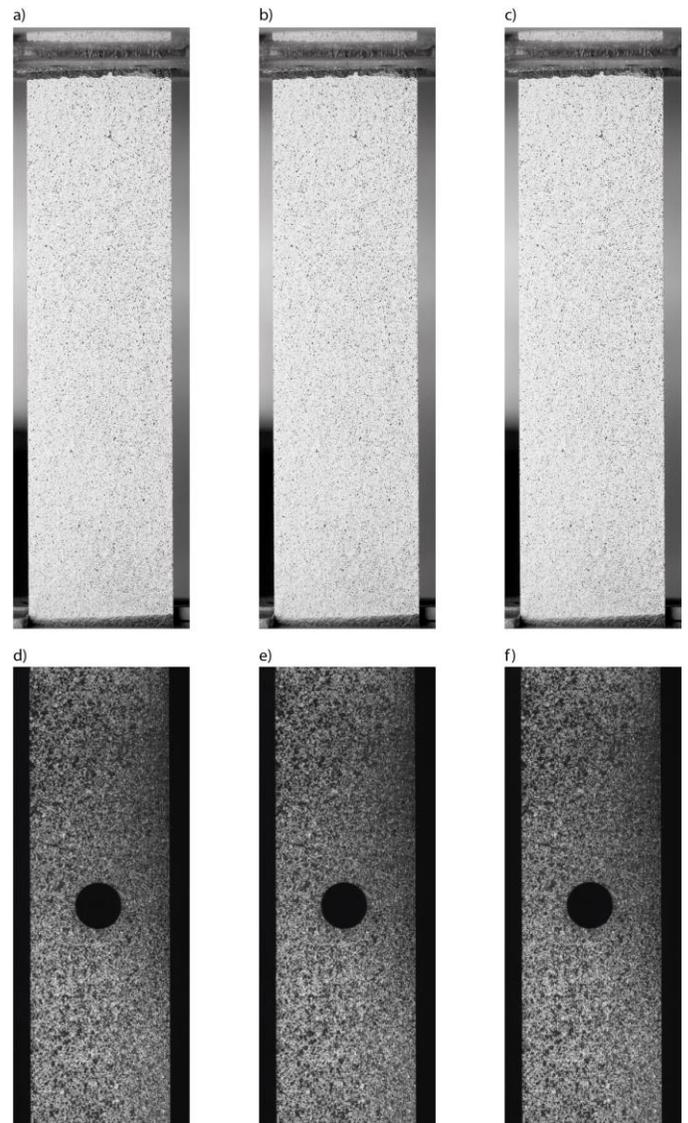

Fig 1: Top: Images of dataset 1 – 0001.jpg (a), 0002.jpg (b) and 0003.jpg (c) ; Bottom: Image of dataset 2 - ohcfrtp_00.tif (d), ohcrfpt_01.tif € and ohcrfpt_02.tif (f) [26]

To evaluate the performance of the different methods the reference image was updated after every evaluation to maximize workload and taking into account that consecutive images in real-time applications have only small displacements. Therefore, every evaluated image pair consists of the two consecutive images. Consequently, there were no large displacements between two images. Nonetheless, the accuracy of the different methods was examined for constant reference images, which results in larger displacements between two images. All methods were implemented such that they have a guaranteed accuracy of 0.1 pixels with results usually having an error below 0.05 pixels.

All tests were carried out with two different hardware settings, generating directly comparable results while also providing information about the influence of the used system. The two used computer systems are listed in Table 3.

| Parameter | Computer 1 | Computer 2 |
|---|---|---|
| Operating System | Windows Server 2012 R2 64-bit | Windows 7 enterprise |
| RAM | 4 GB DDR2 | 16 GB |
| CPU | Intel Xeon Dual Core CPU E5 2726v3 2.4 GHz | Intel Core i7-6700 Quad-Core 3.4 GHz |
| Hard drive | | Samsung PM871a |
| Ethernet Connection | Intel I219-LM | |
| Data Transfere Rate [GBit/s] | 1 | 4,16 |

**Table 3: Hardware Overview Test Computers**

Computer 1 uses two workers per core while Computer 2 uses one worker per core. Nonetheless, both are working with four workers. To minimize the influence of any external factors like high network utilization, all tests have been carried out several times at different days and different times. All presented results are average values of several test runs. Test runs with long calculation times were performed 25 times because the influence of external short duration factors is diminished by the long simulation time itself. Medium to very short simulations were repeated 100, 250 or 1000 times to decrease the influence of short duration factors according to the overall simulation period.

The efficiency evaluation is done for the whole process and not only for the image analysis itself. This includes the program initiation, loading of the images and displaying of results. All computational times presented in Table 4 to Table 6 are the time required from program start until presentation of the last image pairs result. This is contrary to most literature, in which solely the analysis of the image pairs is considered and effects linked to pre- and post-processing are not considered at all.

# 4 Results
## 4.1 Evaluating result correctness
The results of our testing platform were validated by a comparison to results from NCORR by Blaber et al. [26] as shown in Fig 3 and Fig 4.

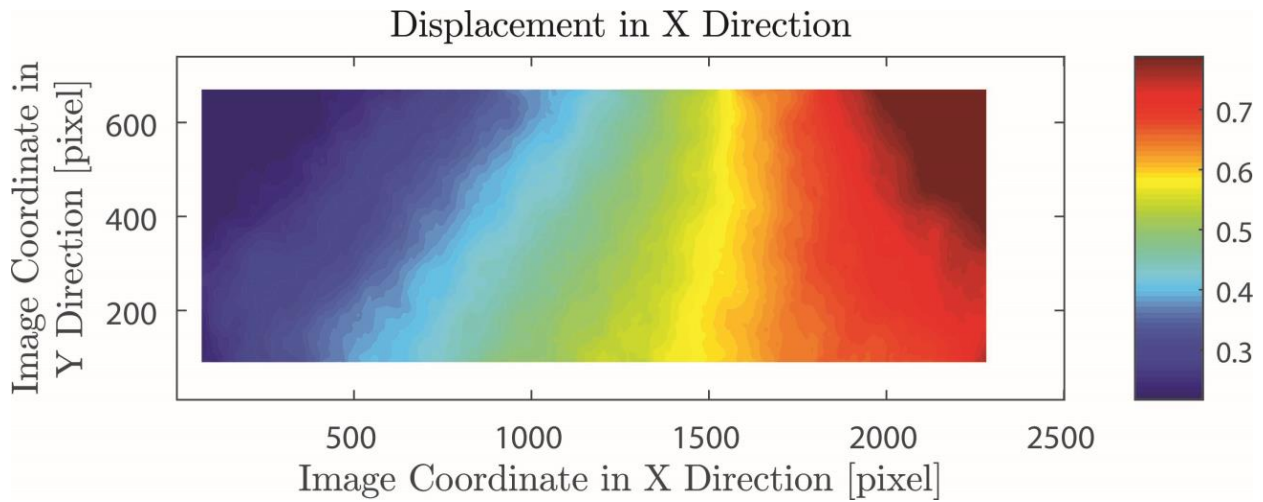

**Fig 2u: Displacement between the first and the 22nd image of data set 1 in horizontal (X) direction. The upper part displays the displacement determined by the authors program.**

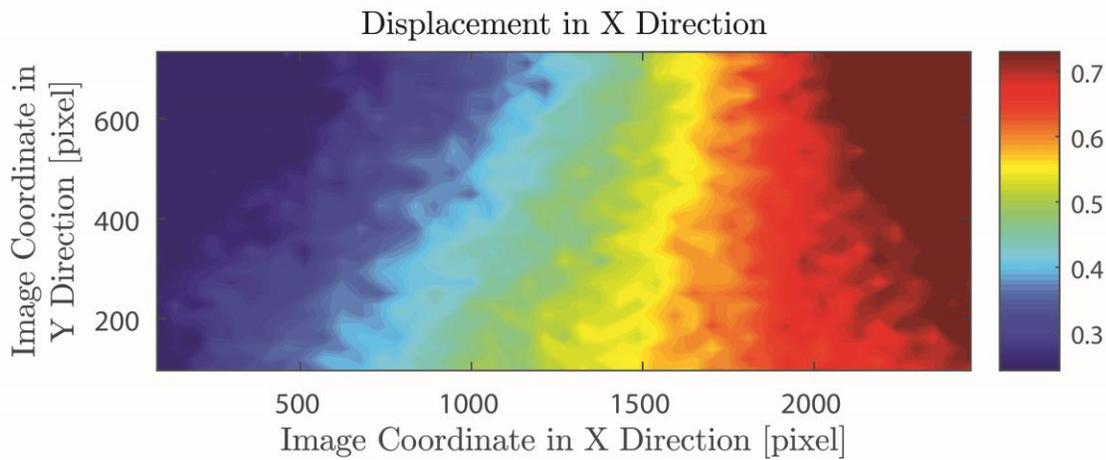

Fig 3l: Displacement between the first and the 22nd image of data set 1 in horizontal (X) direction. The lower part represents the displacement calculated with NCORR developed by Blaber et al. [26].

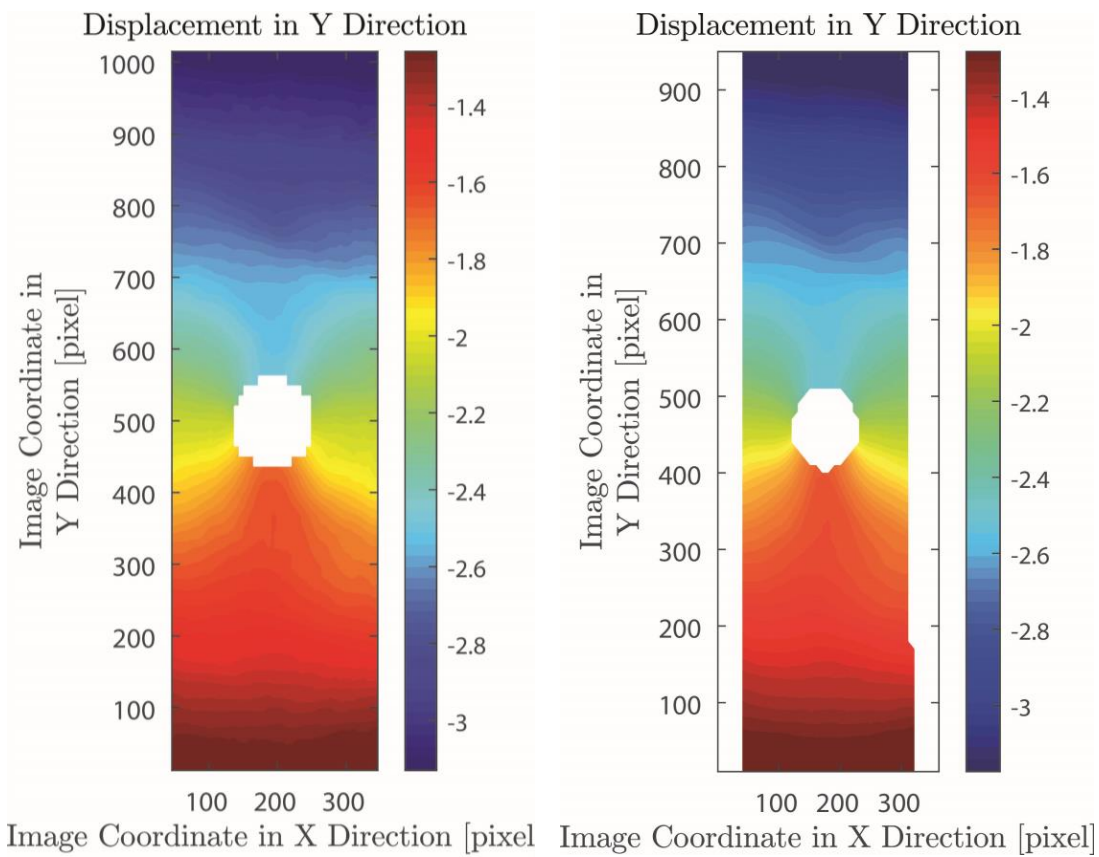

Fig 4: On the left side is the displacement in Y-direction between image 1 and image 4 of data set 2 obtained by the program developed by the authors is shown. On the right site, the same displacement for the same image pair is shown but obtained by the program by Blaber et al. [26].

Comparison of the results obtained by program of the authors and NCORR shows clear similarities. The visualization of the authors' results is slightly different to the visualization of NCORR. Investigation of X- and Y-axis shows that the same section is displayed; investigation of the legend explains the slight differences in the color distribution between the analyses images. This especially holds true for Fig 3 where the upper legends maximum is approximately 0.05 higher than the lower legends one.

## 4.2 Evaluating integer pixel routines

| Integer-Pixel Search algorithm | Sub-Pixel Search algorithm | Time PC1 [s] Set 1 | Time PC1 [s] Set 2 | Time PC2 [s] Set 1 | Time PC2 [s] Set 2 |
|---|---|---|---|---|---|
| Brute Force | Newton-Raphson | **3345.5** | 65.23 | **2084.4** | 45.38 |
| PSO | Newton-Raphson | 582.4 | 19.10 | 346.6 | 7.64 |
| Mod. PSO | Newton-Raphson | 556.9 | *17.94* | 327.2 | *7.34* |
| Brute Force | IC-GN | **3351.3** | 89.44 | **2198.5** | 58.04 |
| PSO | IC-GN | 677.0 | 29.72 | 452.5 | 13.60 |
| Mod. PSO | IC-GN | 706.3 | *30.29* | 464.7 | *12.39* |
| Brute Force | IC-GN by Baker [25] | **3197.2** | 78.86 | **2090.3** | 52.65 |
| PSO | IC-GN by Baker [25] | 534.8 | 20.80 | 309.2 | 7.26 |
| Mod. PSO | IC-GN by Baker [25] | 524.2 | *21.39* | 305.7 | *7.15* |

**Table 4: Overview of the analyses time for serial calculation**

As seen in table Table 4, the processing time between the two different data sets are different for the two hardware settings. A representation of Table 4 in form of a diagram is found in the supplementary material (see S 1).

The relative time difference between the two different data sets in serial computation, for analyses conducted on PC1, is significantly higher if the Brute Force Search algorithm is used. For example, the analysis of data set 1 is around 30 times longer than the analysis of data set 2 if the Particle Swarm Optimization and the Newton Raphson Method are used. If the Brute Force Search algorithm is used, instead of the Particle Swarm Optimization, the ratio of the processing time of data set 1 to data set 2 increases to 50. However, the ratio between the time required analyzing data set 1 to the time required to analyze data set 2 remains constant on PC2, independent of the chosen integer pixel search algorithm.

Generally, all computations processed in serial and conducted on the computer with the better hardware setting (PC 2) were 30% - 70% faster than the same analysis conducted on the computer with inferior hardware properties (PC 1). The time difference between PC 1 and PC 2 for the Brute Force Search based analyses is lower than for the other analyses'. For the integer pixel search algorithms in serial computation (Table 4) it can be seen that all combinations relying on the Brute Force Search algorithm are much slower than the other computations. Moreover both Particle Swarm Optimization as well as the modified Particle Swarm Optimization performed nominally equally on both computers. If the modified particle swarm optimization is used in combination with the Newton Raphson method the whole computation is around 5% faster than the standard Particle Swarm Optimization in combination with the Newton Raphson method. However, a combination of modified Particle Swarm Optimization with the IC-GN algorithm by Baker et al. [25] is only 2% faster than the standard Particle Swarm Optimization combined with the same IC-GN algorithm. A combination of the IC-GN algorithm implemented by the author and the modified Particle Swarm Optimization algorithm is even slower than a combination with the standard Particle Swarm Optimization.

## 4.3 Evaluating sub-pixel routines

Comparing the sub-pixel search algorithms in serial computation (see Table 4) it can be seen that nearly all computations, which use the Newton-Raphson method, are the fastest if compared to analyses' using the same integer-pixel Search algorithm. The only exceptions are the combination of the two Particle Swarm Optimizations (PSO) and IC-GN by Baker et al. [25] for PC 2 and data set 2. Where, both combinations with the IC-GN algorithm by Baker et al. [25] are slightly faster than the Newton-Raphson algorithm. Furthermore, the IC-GN algorithm published by Baker et al. [25] is always faster than the IC-GN algorithm implemented by the author. The difference in performance for the IC-GN algorithm by Baker et al. [25] to the algorithm implemented by the authors is small if the Brute Force Search algorithm is used as an integer-pixel search. If the standard Particle Swarm Optimization is used this difference increases and is maximal if the modified PSO is used.

The time required per image differ substantially between the methods, hardware settings and even image sets. The most time required to analyze one image pair in serial computation was with the Brute Force Search algorithm combined with the IC-GN method by the author conducted on PC 1 with images of data set 1, where it took in average 16.5 seconds per image pair. This equals a processing frame rate of 0.06 Hz. On the other hand the least time required to analyze one image pair in serial computation was with the Modified Particle Swarm Optimization and the IC-GN by Baker et al. [25] conducted on PC2 with data set 2. In this analysis 0.71 seconds were required in average to analyze one image pair which equals a frame rate of 1.4 Hz.

| Integer-Pixel Search algorithm | Sub-Pixel Search algorithm | Time PC1 [s] | | Time PC2 [s] | |
|---|---|---|---|---|---|
| | | Set 1 | Set 2 | Set 1 | Set 2 |
| Brute Force | Newton-Raphson | **2682.5** | 58.66 | **1905.4** | 36.722 |
| PSO | Newton-Raphson | 863.5 | 20.00 | 677.6 | 8.38 |
| Mod. PSO | Newton-Raphson | 861.5 | *19.41* | 618.9 | *8.17* |
| Brute Force | IC-GN | **2938.0** | 61.56 | **1942.1** | 38.756 |
| PSO | IC-GN | 914.6 | 24.18 | 678.8 | 11.10 |
| Mod. PSO | IC-GN | 964.6 | *24.15* | 685.7 | *10.99* |
| Brute Force | IC-GN by Baker [25] | **2348.4** | 55.13 | **1776.4** | 35.00 |
| PSO | IC-GN by Baker [25] | 552.2 | 17.98 | 424.2 | 7.49 |
| Mod. PSO | IC-GN by Baker [25] | 902.7 | *20.19* | 418.9 | *6.60* |

Table 5: Overview of the analyses time for parallel Sub-Image calculation

As seen in Table 5, the time differences between the two different data sets are different for the two hardware settings. A visualization of Table 5 can be found in the supplementary materials (see S 2) All computations conducted on the computer with the better hardware setting (PC 2) are 30 - 200 % faster than the same analysis conducted on the computer with inferior hardware setting (PC 1) if processed in sub-image parallel computation. Here it can be seen that the time difference between PC 1 and PC 2 tends to be smaller for the Brute Force based analyses than for the other analyses. Additionally, the influence of hardware settings is stronger for data set 2 than for data set 1. Furthermore, the time required to analyze the two different data sets differs more on PC 2. The difference is small if the Brute Force Search algorithm is used but the difference is larger if the Particle Swarm Optimization or the modified Particle Swarm Optimization is used.

### 4.4 Significance of parallel computation

Looking at the integer pixel search algorithms in sub-image parallel computation it can be seen that all the combinations relying on the Brute Force Search algorithm are much slower than the other computations, Table 5. The Brute Force search based computations are between three to five times slower than the other algorithms. Moreover, it can be seen that both, the Particle Swarm Optimization as well as the modified Particle Swarm Optimization, perform more or less equal. In combination with the Newton Raphson method is the modification around 5 % faster than the standard Particle Swarm Optimization.

In the case of sub-pixel search algorithms in parallel sub-image computation, in Table 5 it can be seen that computations using the IC-GN algorithm implemented by the authors is always slower than any of the other two. The time required per image is different between the methods, hardware settings and even image sets. The most time required to analyze one image pair in parallel sub-image computation was with the Brute Force Search algorithm with IC-GN method by the author conducted on PC 1 with images of data set 1, where it took in average 14.47 second per image pair. This equals a frame rate of 0.07 Hz. The least time required to analyze one image pair in parallel sub-image computation was with the Modified Particle Swarm Optimization and the IC-GN by Baker et al. [25] conducted on PC2 with data set 2. In this analysis 0.66 seconds were required in average to analyze one image pair which equals a frame rate of 1.5 Hz.

| Integer-Pixel Search algorithm | Sub-Pixel Search algorithm | Time PC1 [s] | | Time PC2 [s] | |
|---|---|---|---|---|---|
| | | Set 1 | Set 2 | Set 1 | Set 2 |
| Brute Force | Newton-Raphson | **1778.7** | 41.20 | **1208.3** | 24.94 |
| PSO | Newton-Raphson | 152.8 | 5.57 | 105.5 | 2.90 |
| Mod. PSO | Newton-Raphson | 158.9 | *5.44* | 108.0 | *2.89* |
| Brute Force | IC-GN | **1823.6** | 45.66 | **1239.8** | 26.13 |
| PSO | IC-GN | 186.9 | 8.82 | 124.4 | 4.36 |
| Mod. PSO | IC-GN | 189.7 | *8.10* | 135.8 | *4.36* |
| Brute Force | IC-GN by Baker [25] | **1766.4** | 43.32 | **1193.6** | 24.73 |
| PSO | IC-GN by Baker [25] | 156.0 | 5.63 | 107.2 | 2.99 |
| Mod. PSO | IC-GN by Baker [25] | 148.8 | *5.74* | 106.0 | *2.91* |

**Table 6: Overview of the analyses time for parallel Image calculation**

As seen in Table 6, the time differences between the two different data sets are different for the two hardware settings. A visualization of Table 6 can be found in the supplementary materials (see S 3). All computations conducted on the computer with the better hardware setting (PC 2) are 40 - 100 % faster than the same analyses conducted on the computer with inferior hardware setting (PC 1), if they are processed in image parallel computation. The time difference between PC 1 and PC 2 tends to be smaller for the Brute Force Search based analyses than for the other analyses if dataset 2 is analyzed while there is not such an influence for data set 1. Additionally, the influence of the hardware setting is stronger for data set 2 than for data set 1. Furthermore, the time required to analyze the two different data sets differs more on PC 2. The difference is small if the Brute Force Search algorithm is used but big for the other two search algorithms. For example, the analysis of Set 1 took 43 times as long as the analysis of Set 2 with a combination of PSO and NR algorithm on PC 1, while the same analysis took 80 times longer for Set 1 than for Set 2 on PC 2.

Regarding the integer pixel search algorithms in parallel image computation (see Table 6) it can be seen that all combinations relying on the Brute Force Search algorithm are much slower than the other computations. The Brute Force search based computations are between three to four times slower than the other algorithms. Moreover, the Particle Swarm Optimization as well as the modified Particle Swarm Optimization, performed in the same order of magnitude. A combination of Newton Raphson method and modified Particle Swarm Optimization is 5% faster than a combination of Newton Raphson method and standard Particle Swarm Optimization. A combination of the modified Particle Swarm Optimization and the IC-GN algorithm by the authors performs equally to a combination of the standard Particle Swarm Optimization and the same IC-GN algorithm. However, if the IC-GN algorithm published by Baker et al. [25] is used instead, the variations in performance cannot be generalized, as can be seen in Table 6.

For the sub-pixel search algorithms in parallel image computation (see Table 6) it can be seen that computations which use the Newton-Raphson method are faster than computations using the IC-GN algorithm implemented by the author, while computations relying on the IC-GN algorithm published by Baker et. al [25] are faster. Furthermore, the Inverse-Compositional Gauss-Newton (IC-GN) algorithm published by Baker et al. [25] is always faster as the IC-GN algorithm as implemented by the author. The most time required to analyze one image pair in parallel image computation was with the Brute Force Search algorithm combined with IC-GN method by the author conducted on PC 1 with images of data set 1, where it took in average 8.98 second per image pair. On the other hand the least time required to analyze one image pair in parallel image computation was with the Modified Particle Swarm Optimization and the Newton-Raphson method conducted on PC2 with data set 2. In this analysis 0.29 seconds were required in average to analyze one image pair which equals a frame rate of 3.5 Hz. The same frame rate could be achieved with a combination of the modified Particle Swarm Optimization and the Newton-Raphson method as well as with the modified Particle Swarm Optimization and the IC-GN by Baker et al. [25].

Comparing Table 4 to Table 6 shows that parallel image computation is more efficient by being two to three times faster compared to serial computation. Furthermore, it is remarkable that the time difference is smaller for method combinations relying on the Brute Force Search algorithm. Finally, parallel sub-pixel calculation does not necessarily improve the computational speed of a serial computation.

Comparing Table 4 and Table 5 shows that parallel sub-image computation is inferior to serial computation in most cases. The parallelization is only beneficial if the Brute Force Search algorithm is used as an integer pixel search algorithm. In general, the larger images of data set 2 benefit in parallel computation more than the smaller images of data set 1.

# 5 Discussion
## 5.1 Hardware profile and influence of data set properties

All analyses executed on PC 2 are faster than the same analyses conducted on PC 1. This is an expected outcome as well, because PC 2 has a far better hardware setting than PC 1. One of the most important hardware differences between the two Computers is the CPU. PC 1 has a 3.4 GHz Quad-Core CPU while PC 1 just has a 2.4 GHZ Dual Core CPU. Therefore, PC 2 has 1.4 times better kernels in terms of frequency than Computer 1. This relation is observable in the relationships of the analyses times as well. Nonetheless, other factors as cache size and data bus speed also have an influence. The data transfer rate between image storage place and CPU is also a limiting factor. For PC 1 this is, most likely, the Ethernet connection that connects the system to the central server. This Ethernet connection has a data transfer rate of 1 GBit/s. Additionally, also several other data has to be transferred via this Ethernet connection. Therefore, the whole bandwidth is not available for the program itself. PC 2 has to read the images from its SATA SSD, which has a reading speed of 4.16 GBit/s. It is assumed that no other data transfers are performed during DIC runs; therefore, the entire transfer rate is available for the program. These two factors influence the time required to read in one image pair and directly leads to a worse performance on PC 1. Techniques as double buffering are not considered. It is assumed that the program always uses the most current image of a live video stream as target image, with the previously used target image as new reference image. This is done to avoid queuing of not processed images if the camera acquires images faster than the program processes them.

Furthermore, the improvements between the two different computer settings decrease with increasing data traffic. This is explained by the rising portion of data handling time in the total computation time with increasing data traffic. The data transfer speed between the different workers can be assumed to be about the same for the different computers and, therefore, the time required to transfer a specific amount of data is the same for the superior hardware setting as for the inferior one.

The processing times also differ between the two different data sets. The influence of the Brute Force search algorithm is stronger for data set 1 than for data set 2 if comparing this integer pixel search algorithm to the others. The Brute Force Search algorithm determines the correlation coefficient distribution for the whole image. Therefore, the computational time of one sub-image is directly related to the image size. The both Particle Swarm Optimizations work always with a fixed number of particles and generations. Therefore, the computational time of those two algorithms is not directly related to the image size itself. In conclusion, the computation time rises with increasing image size for the Brute Force Search while it remains constant for the two variations of the Particle Swarm Optimization. Of course this only holds true if a constant number of sub-images is chosen.

## 5.2 Integer pixel search

For the integer-pixel search algorithms, in general, the Brute Force Search algorithm is by far the most inefficient algorithm. This was excepted because this very simple algorithm determines the correlation coefficient for every possible sub-pixel position, which requires a significantly higher computational effort than the two particle Swarm Optimizations. This is explained by the following calculation. Assuming that the images have a size of 500 x 1000 pixels and the sub-sets have a size of 31x 31 pixels. This gives a total of 970 x 470 = 455,900 possible sub-set positions within one image. Consequently, for the Brute Force Search algorithm 455,900 correlation coefficients have to be determined per reference sub-image. Contrary, both Particle Swarm Optimizations use a total number of 50 particles over a maximum of five iterations. In the standard particle Swarm Optimization the correlation coefficient of each particle for each iteration step is determined, resulting in maximal 250 evaluations of the correlation coefficient. For the modified Particle Swarm Optimization the correlation coefficient of the particle as well as the four surrounding pixel is determined which results in a theoretical total of maximal 1250 evaluations of the correlation coefficient. Both algorithms use look-up tables to reduce the computational burden if different particles move to the same position. Additionally, the modified Particle Swarm Optimization converges faster due to its "smarter" particles leading to a lower number of iteration steps in average. Besides the determination of the correlation coefficients for the different positions, all three methods determine the maximal found correlation coefficients. However, this can be considered less computational intense as the determination of the correlation coefficient itself. By simply comparing the number of calculated correlation coefficients of the different methods it becomes obvious that the Brute Force Search algorithm requires far more computation power than the two Particle Swarm Optimizations.

Unfortunately, the case is not that clear for the standard Particle Swarm Optimization (PSO) and the modified version. The significant difference between those two methods is that the standard version only takes the correlation coefficient at the particles own position into account while the modified version additionally takes also the correlation coefficients at the surrounding pixels into account. This leads to a faster convergence because more information are taken into account while updating the particle positions. This also increases the total computational effort per iteration step. Because the Particle Swarm Optimization stops if a sufficiently high correlation coefficient of 0.995 is found, the number of used generations is lower for the modified Particle Swarm Optimization than for the standard version. With respect to the results gotten in the conducted tests and presented in Table 4 to Table 6, it can be stated that the modified PSO version performs in general at least equally well as the standard version. Therefore, in most cases there will be no significant disadvantage if the modified version is used but a benefit might be possible.

### 5.3 Sub-pixel search

The version of the Inverse-Compositional Gauss-Newton algorithm as published by Baker et al. [25] is around 2-3% more efficient than the version implemented by the authors. The reasons for this discrepancy can be many and very detailed in the specific implemented code. One major difference between the two versions is the data flow to sub-functions. The version published by Baker et al. [25] uses less sub-functions than the version implemented by the author. Otherwise, the amount of data passed by the authors implementation is lower per sub-function call than the amount of data passed in the version by Baker et al. [25].

The performance comparison between the Newton-Raphson method and the Inverse-Compositional Gauss-Newton algorithm by Baker [25] cannot be generalized and apparently both algorithms perform similarly as can be seen in Table 4 to Table 6. Among all tests conducted as part of this study sometimes, the Newton-Raphson algorithm is faster, sometimes, the IC-GN algorithm by Baker et al. [25] is faster and sometimes there is no difference. This somewhat contradicts theoretical predictions and most current literature, which clearly states that the IC-GN algorithm is superior to the Newton-Raphson algorithm [18]. The implementation of the IC-GN algorithm into the program was more complex compared to the NR algorithm. The fact that Matlab generally uses call by value functionality strongly increases the time required to shift data from different sub-functions. Every time a sub-function is called, a copy of the data, which has to be transferred between caller function and called function, is created. This drastically increases the computation time. Because the data flow is more complex for both IC-GN versions than for the NR algorithm, Matlab requires more time to copy values and allocate memory. This diminishes the advantages of the IC-GN compared to the NR algorithm. Furthermore, the chosen accuracy is such that often a small number of iterations is required. However, the IC-GN algorithms advantage is the improved calculation time per iteration step. If the number of iterations is small this advantage is negligible. The overall influence of this is only hardly measurable on a theoretical basis. The fact that there is no real evidence that one algorithm is more efficient than the other (except for the IC-GN by the author) but a slight connection to the data sets supports the premise that the reason for the worse performance than theoretically excepted is mainly connected to the data flow and the small number of iterations.

### 5.4 Parallel computation

With respect to the different computing types, the conducted tests clearly show that parallel image computation is far more efficient than serial and parallel sub-image computation; compare Table 4 to Table 6. Parallel image computation basically solves the same task as in serial computation. In both cases, the worker solves one image completely on its own. In serial computation a single worker analyses all image pairs, while in parallel image computation several workers are working on different image pairs at the same time. Nonetheless, the data has to be organized and distributed to the workers, which is more complex in the case of parallel image computation creating an additional workload for this computational type. However, this workload is, compared to the analysis itself, small. Otherwise, parallel sub-image computation differs more from serial computation. In parallel sub-image computation the different workers have to solve different subsets of the same image pair. To do so, all workers need both, the target as well as the reference image. Consequently, this increases the data flow because every worker needs every image pair and slows down the computation. Assuming a number of four workers, which is the case for both hardware settings used in this study, this results in a four times bigger data amount for sub-image parallel computation than for image parallel computation. Therefore, the overall efficiency of sub-image parallel computation is lower than parallel image computation. Nonetheless, all workers are working on the same image pair in sub-image parallel computation. This means that the time between analysis start of a specific image pair and the determination of the final displacement result is shorter than for image parallel computation.

handle the additional data traffic is higher than the time reduction through the task parallelization. Nonetheless, the conducted tests show that the correct parallelization type can reduce the computation time to up to 55%, even on a standard quad-core processor. Finally, the tests show that theoretically superior methods can perform worse than other methods because of some practical implementation disadvantages of these methods.

# 6 Conclusion

The analysis of the different mathematical search algorithms shows that there are substantial differences between the tested methods and combinations regarding computational time despite similar accuracies. It is clear that a combination of highly efficient methods can reduce the computational time of standard Digital Image Correlation while maintaining the same accuracy. Furthermore, the conducted tests reveal that a real-time analysis with a high frame rate requires computational resources that are only present in high-end computing systems; or are only possible at the expense of significant reduction of accuracy. Nonetheless, real-time analysis with a slow frame rate, lower accuracy or a reduced number of evaluated data points is possible in MATLAB with nominally average computing systems - analysis frequencies up to 3.5 Hz are possible with a standard Intel i7 Quad-Core CPU.

The fastest computation can be achieved by a combination of a modified Particle Swarm Optimization, in which a Downhill Star Search algorithm is integrated into the particle swarm optimization search algorithm, as an integer pixel search algorithm. To increase the accuracy to subpixel level a Newton-Raphson Search algorithm is best suited when a relatively low accuracy of 0.1 pixels is required. The computational time is least if image pairs are processed in parallel. Parallelization of smaller tasks, as in sub-image parallel computation, lead to computational times higher than in serial processing. The time required to

# Supplementary Material

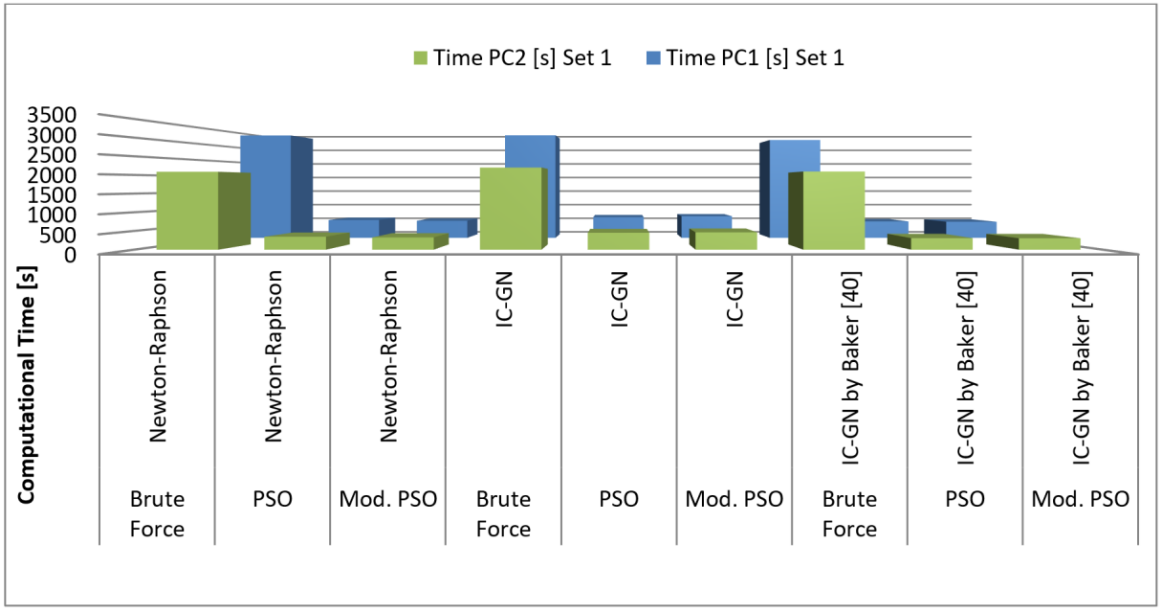

S 1: Overview of the Total Computational Time of data set 1 in serial computation

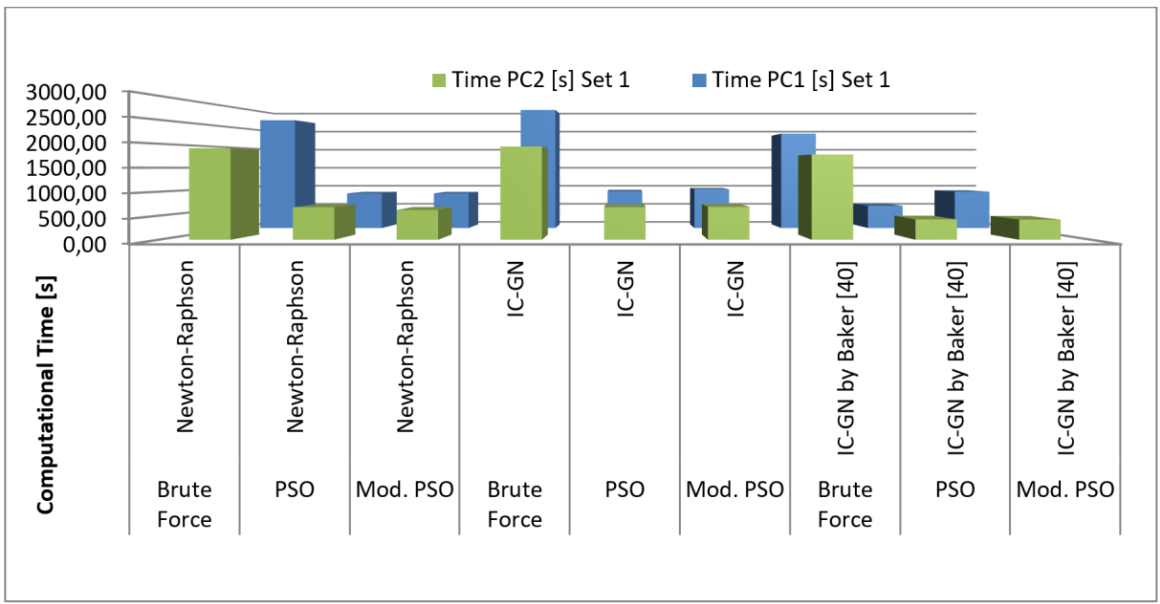

S 2: Overview of the Total Computational Time of data set 1 in parallel sub-image computation

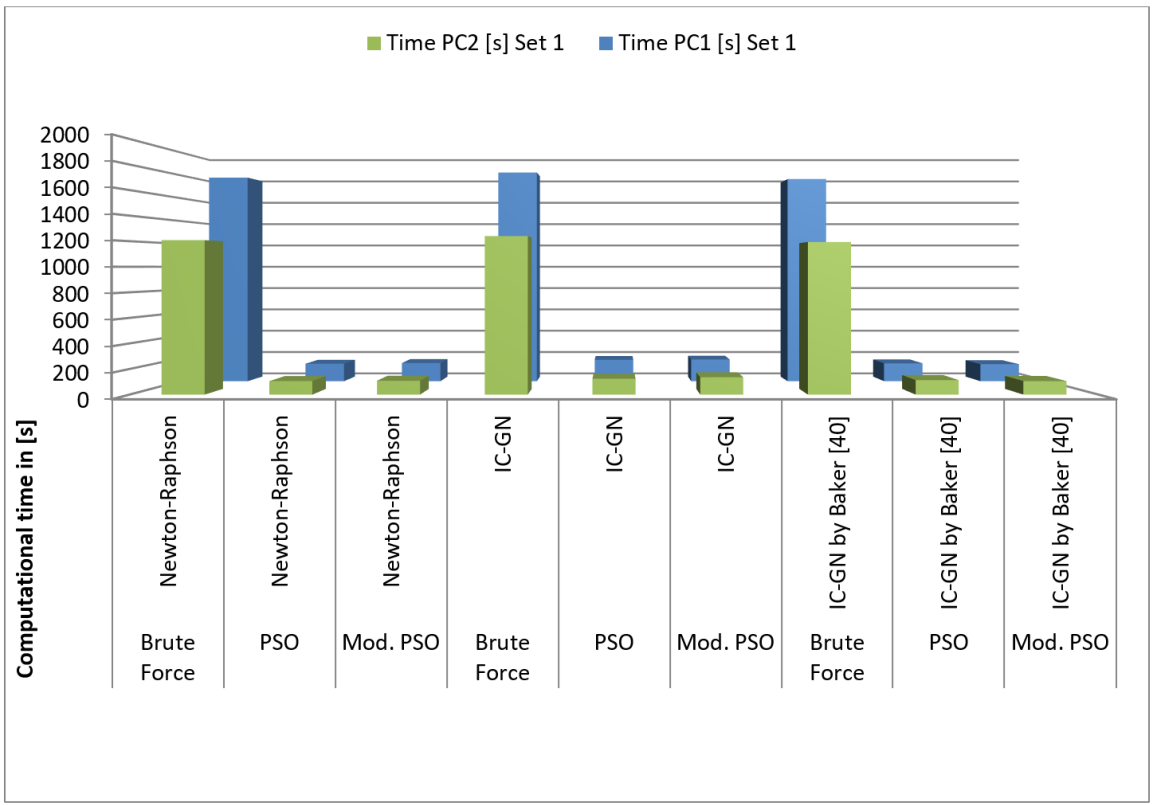

S 3: **Overview of the Total Computational Time of data set 1 in image parallel computation**